\documentclass[sigconf]{acmart}

\usepackage[caption=false,font=footnotesize]{subfig}

\renewcommand\footnotetextcopyrightpermission[1]{} 
\usepackage{booktabs} 

\setcopyright{none}

\begin{document}
\title{Developing All-Skyrmion Spiking Neural Network}

\author{Zhezhi He and Deliang Fan}
\orcid{1234-5678-9012}
\affiliation{%
  \institution{Dept. of Electrical and Computer Engineering, University of Central Florida, Orlando, Florida 32816}
}
\email{Elliot.He@Knights.ucf.edu, Dfan@ucf.edu}

\begin{abstract}
In this work, we have proposed a revolutionary neuromorphic computing methodology to implement \textit{All-Skyrmion Spiking Neural Network} (AS-SNN). Such proposed methodology is based on our finding that skyrmion is a topological stable spin texture and its spatiotemporal motion along the magnetic nano-track intuitively interprets the pulse signal transmission between two interconnected neurons. In such design, spike train in SNN could be encoded as particle-like skyrmion train and further processed by the proposed skyrmion-synapse and skyrmion-neuron within the same magnetic nano-track to generate output skyrmion as post-spike. Then, both pre-neuron spikes and post-neuron spikes are encoded as particle-like skyrmions without conversion between charge and spin signals, which fundamentally differentiates our proposed design from other hybrid Spin-CMOS designs. The system level simulation shows 87.1\% inference accuracy for handwritten digit recognition task, while the energy dissipation is $\sim$ 1 fJ/per spike which is 3 orders smaller in comparison with CMOS based IBM TrueNorth system.
\end{abstract}


\maketitle

\section{Introduction}
State-of-the-art Complementary Metal-Oxide Semiconductor (CMOS) Boolean logic and Von Neumann architecture based computers are vastly less energy efficient and slower than human brains at the same size in various cognitive applications, such as visual/speech recognition, inference and intelligent data/signal processing. There is a great need to build a neuromorphic computing platform capable of performing human-like cognitive computing in energy efficient manner. Spiking Neural Network (SNN) is a promising neuromorphic computing model for the upcoming data explosion due to its unsupervised learning ability with unlabelled data. With neuron and synapse as two essential computing units, SNN constructs a multiple neuron layer topology, where neurons are interconnected with programmable synapses. Due to the biological plausible operating mechanism, both inputs and outputs of SNN are encoded as spike trains. 

\begin{figure}[h]
	\centering
	\captionsetup[subfloat]{farskip=2pt,captionskip=1pt}	
	\includegraphics[width=0.49\textwidth]{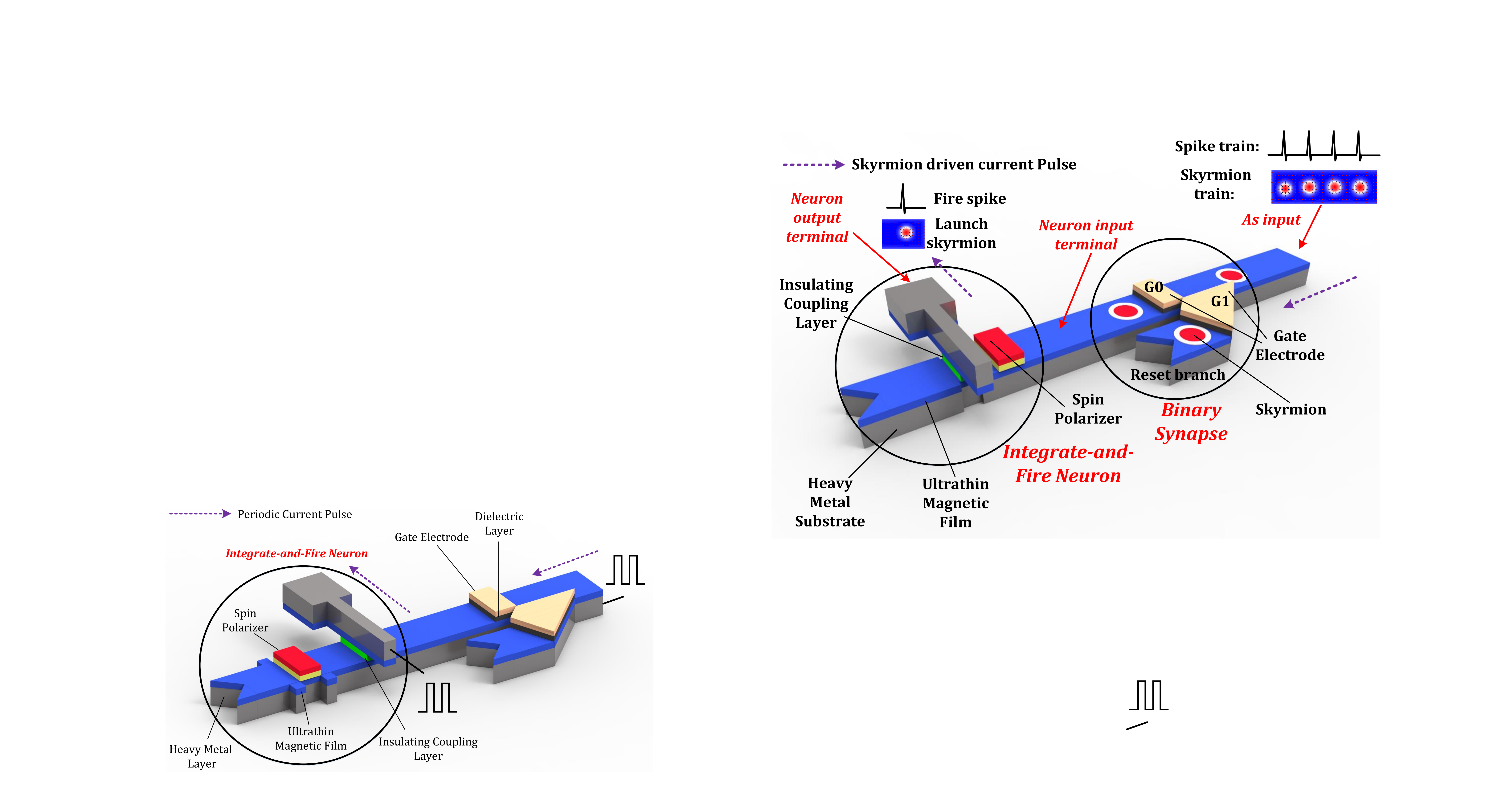}
	\caption{Structure of proposed all-skyrmion elementary cell, consisting of binary synapse and integrate-and-fire neuron. Note that, only one synapse is shown for simplicity.}
	\label{Device}
\end{figure}

Magnetic skyrmion is an ultra-small particle-like spin texture, which can stably exist in heterogeneous device structure (ultra-thin magnetic film with heavy metal substrate in Fig.~\ref{Device}) \cite{fert2013skyrmions}. Magnetic skyrmion, as compared with traditional spintronic devices (e.g. Magnetic Tunnel Junction (MTJ), Domain Wall Motion (DWM) devices), has several unique device properties including ultra-small size (sub-10nm), high current-driven skyrmion drifting velocity ($\sim$75m/s) with ultra-low depinning current density (10\textsuperscript{8} A/m\textsuperscript{2}, 2-4 orders lower than DWM devices) and high defect tolerance \cite{jiang2015blowing}. The generic application of skyrmion is used as information carrier, where binary data (`1' or `0') is encoded by the presence or absence of skyrmion.

In this work, we propose a revolutionary methodology to implement SNN, called \textit{All-Skyrmion Spiking Neural Network} (AS-SNN). Such proposed methodology is based on our finding that skyrmion is a topological stable spin texture and its spatiotemporal motion along the magnetic nano-track intuitively interprets the pulse signal transmission between two interconnected neurons (Fig.~\ref{Device}). In such design, spike train in SNN could be encoded as particle-like skyrmion train and further processed by the proposed skyrmion-synapse and skyrmion-neuron within the same magnetic nano-track to generate output skyrmion as post-spike. Then, both pre-neuron spikes and post-neuron spikes are encoded as particle-like skyrmions without conversion between charge and spin signals, which fundamentally differentiates our proposed design from other hybrid Spin-CMOS designs \cite{srinivasan2016magnetic,sengupta2016hybrid}. Such all-skyrmion methodology mimics biological neural network and synapse/neuron are automatically implemented by the intrinsic device physics without complex peripheral control circuits, leading to potential three orders higher energy efficiency of CMOS counterparts.

\section{All-Skyrmion SNN}

In this section, we simply introduce the neuron and synapse design of proposed all-skyrmion spiking neural network structure and methodology. 


\subsection{Binary skyrmion synapse}
In a standard SNN, synapse is used to store weight parameters and multiple the stored weight with input spike signals. Typically, the distribution of normalized synaptic weights of entire SNN reveals that, the majority of synaptic weights are located near `0' (i.e. eliminate the input spike) or `1' (i.e. pass the input spike without modulation). Such synaptic weights distribution indicates the potential of weight binarization with acceptable inference accuracy degradation. Thus in this work, we have employed the binary synapse weight as shown in Fig.~\ref{Device}, where the skyrmion based binary synapse is realized through a 'y-junction' nano-track geometry with electrode pinning sites to select the skyrmion motion path. Through utilizing the Voltage Control Magnetic Anisotropy (VCMA) effect \cite{kang2016complementary}, two electrode gates (G0 and G1) are applied with complementary voltage signals working as synaptic weight. In this case, when the synaptic weight is `1' (G0=0V, G1=5V), the input skyrmion will only pass G0 to the neuron region since a magnetic anisotropy barrier or well is created under G1 due to VCMA. Similarly, if weight is '0' (G0=5V, G1=0V), the input skyrmion will be directed to the reset branch for annihilation. Equivalently, the input spike signal (i.e. skyrmion) are eliminated due to weight='0'. Note that, this work only discusses constructing all-skyrmion SNN system with focus of skyrmion-synapse and -neuron design. For learning of AS-SNN, additional non-volatile bit-cell and similar corresponding control circuit in \cite{srinivasan2016magnetic} could be used to realize unsupervised learning.

\subsection{Skyrmion neuron}

\begin{figure}[h]
	\centering
	\captionsetup[subfloat]{farskip=2pt,captionskip=1pt}	
	\subfloat[\label{neuron1}]{%
		\begin{minipage}[c][0.85\width]{0.28\textwidth} 
			\centering
			\includegraphics[width=\textwidth]{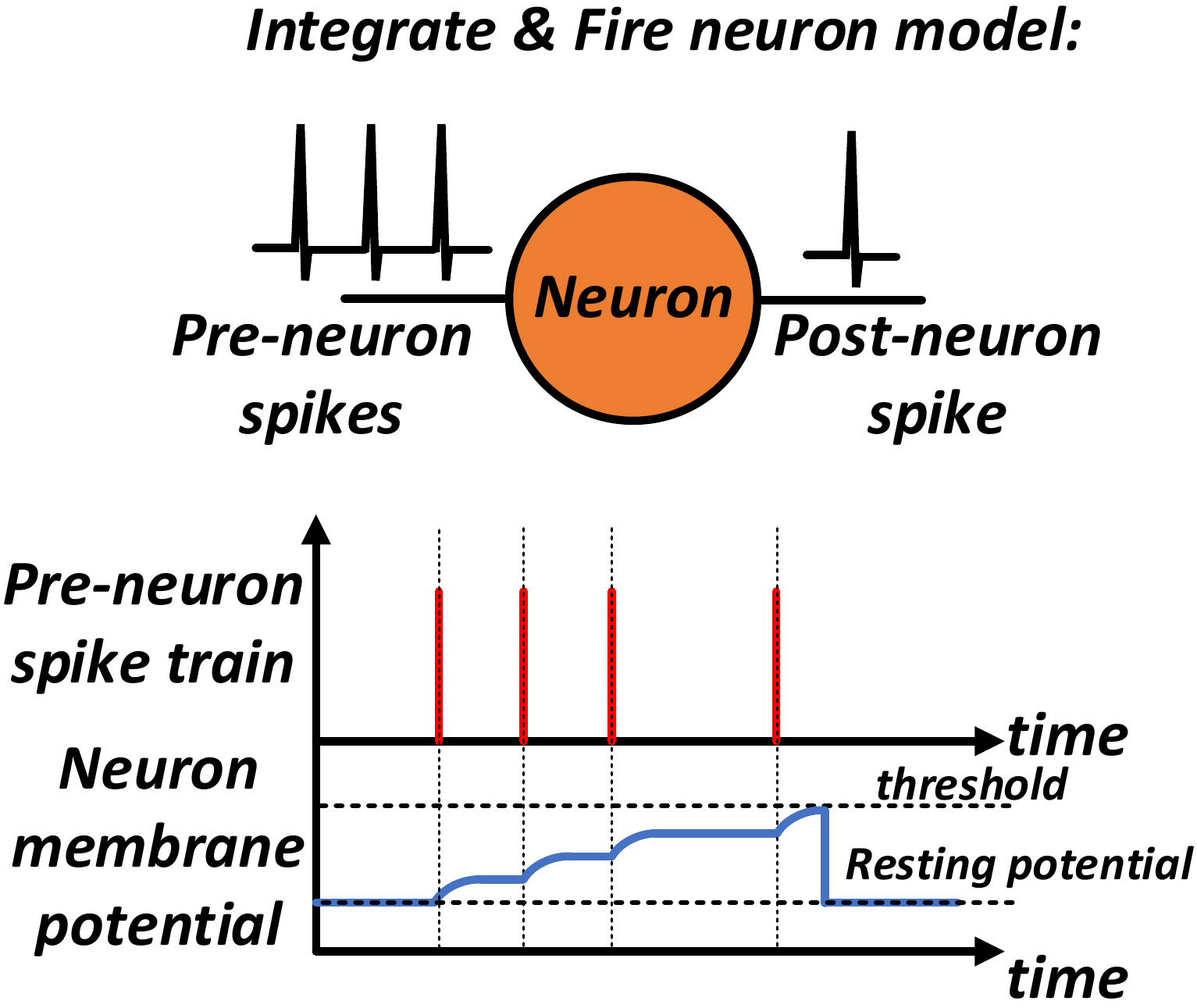} \end{minipage}}\hfill
	\subfloat[\label{neuron2}]{%
		\begin{minipage}[c][1.4\width]{0.17\textwidth} 
			\centering
			\includegraphics[width=\textwidth]{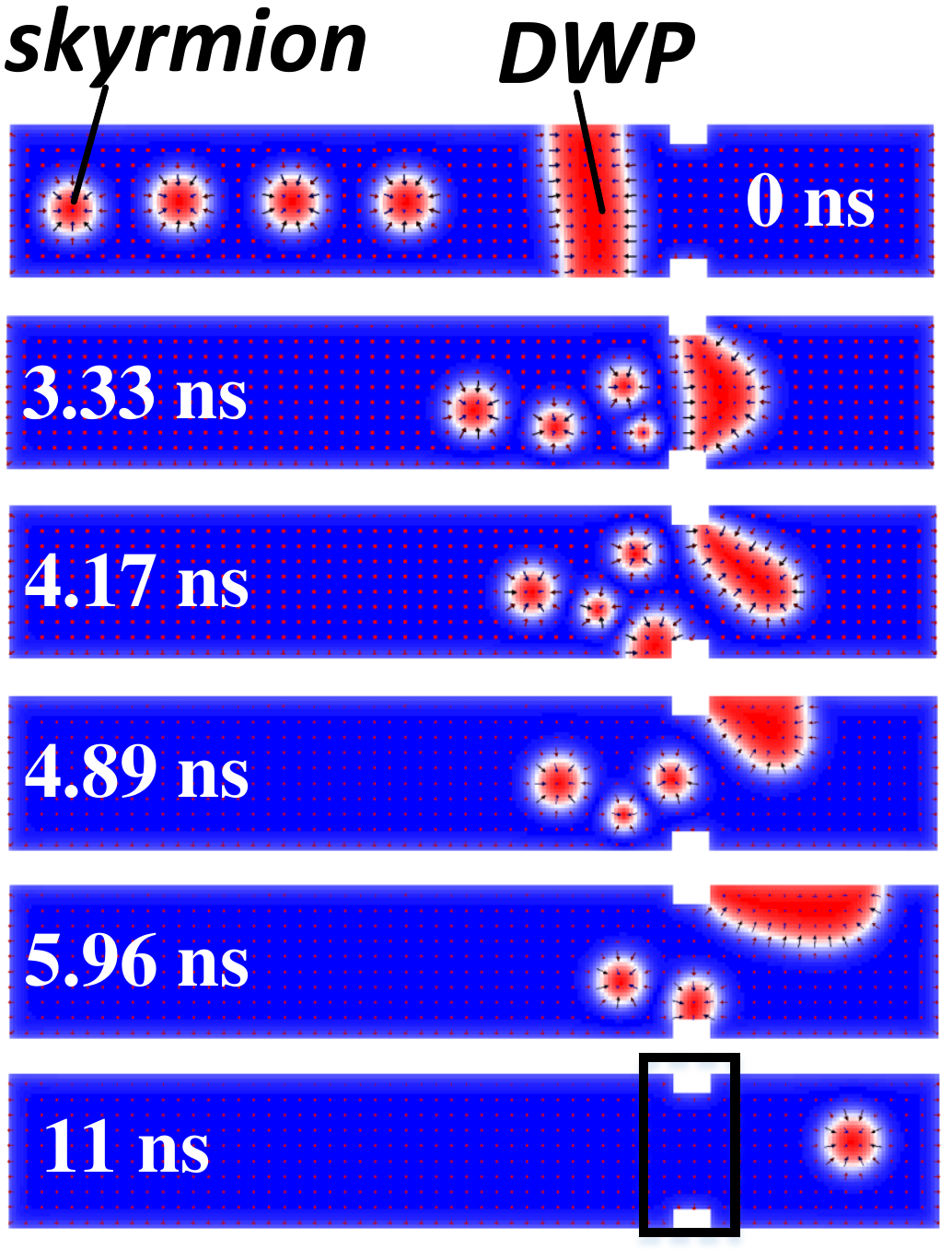} \end{minipage}}
	\caption{(a) The integrate-and-fire neuron model. (b) micromagnetic dynamics of skyrmion-based neuron design.} 
	\label{neuron}
\end{figure}

In AS-SNN design, We choose the classical Integrate-and-Fire (IF) neuron as the neuron model. In typical IF neuron, it takes biological electrical pulse signals and accumulates on the membrane, then fires a pulse when the membrane potential reaches its threshold as shown in Fig.~\ref{neuron1}. Here, we propose a skyrmion-neuron design, which performs similar IF function with skyrmions. As device structure of skyrmion neuron depicted in Fig.~\ref{Device}, it integrates the skyrmions modulated by the incoming binary synapse from the neuron input terminal, and fires a skyrmion at the neuron output terminal when the number of skyrmion within the neuron region reaches the preset threshold. 

\textbf{\underline{Integrate:}} In our recent work \cite{he2017current}, we have reported the current-driven dynamics between multiple skyrmions and Domain Wall Pair (DWP) within the skyrmion nano-track. The DWP (generated by the spin polarizer) that is trapped within the notch region of nano-track, can be depinned by the accumulation of specific number of skyrmions. Note that, such number could be adjusted through tuning driving current, pinning site geometry, etc. The micromagnetic dynamics of DWP depinned by four skyrmion (i.e. neuron threshold) is shown as an example in Fig.~\ref{neuron2}. Such domain wall pair depinning behavior can be directly leveraged to mimic skyrmion-neuron integration and fire function.

\textbf{\underline{Fire:}} In order to realize all-skyrmion based neuron design (i.e. both inputs and outputs are skyrmion), we take advantage of magnetic coupling effect \cite{morris2012mlogic} and propose a novel nanotrack geometry as shwon in Fig.~\ref{Device}. Thus, when the domain wall pair is depinned by the input skyrmions (number of input skyrmions reaching the threshold), another domain wall pair will be nucleated due to magnetic coupling in the adjacent neuron output terminal, which is isolated by the insulating coupling layer. Then such newly formed DWP could be converted into an output skyrmion due to specially designed device geometry, which has been proven in \cite{zhou2014reversible}. Above all, we could build a skyrmion-neuron which integrates input skyrmion train and fires a new skyrmion if reaching a preset threshold. 

\textbf{\underline{Fan-out:}} The drivability of neuron is another major design concern, since each pre-neuron is interconnected with large amount of post-neurons. Because skyrmion is used as the information carrier instead of electrical pulse, energy-efficient duplication of Skyrmion is a possible solution to increase fan-out. As proven in \cite{zhang2015magnetic}, skyrmion duplication could be performed during its propagation within the Y-junction nano-track geometry, which is similar as cell division. 

\textbf{\underline{Reset:}} Due to the event-driven characteristic of our skyrmion neuron, the only reset operation required is to nucleate a new domain wall pair by spin polarizer (Fig.~\ref{Device}) after neuron fires (i.e. during the refractory period).

\section{Performance evaluation}
\textbf{\underline{Device:}} In conclusion, the proposed All-Skyrmion Spiking Neural Network (AS-SNN) has following primary advantages: 1) There is no signal conversion between Skyrmion and electrical pulse, since skyrmion is not only used as information carrier for signal transmission between neurons, but also employed as the computation medium (i.e. Skyrmions-DWP interaction dynamics \cite{he2017current}, DWP-skyrmion conversion \cite{zhou2014reversible}), thus intrinsically emulating the biological neural network and leading to revolutionary neuromorphic computing methodology. 2) The power consumption of such AS-SNN design primarily comes from skyrmion driven current and neuron reset. Due to the high skyrmion drifting velocity ($\sim$75m/s) with ultra-low driven current density ($\sim6 \times 10^10$ A/m\textsuperscript{2}), the energy dissipation is $\sim$1 fJ/per spike which is more than 3 orders smaller in comparison with IBM TrueNorth \cite{merolla2011digital}.

\begin{figure}[h]
	\centering
	\captionsetup[subfloat]{farskip=2pt,captionskip=1pt}	
	\includegraphics[width=0.40\textwidth]{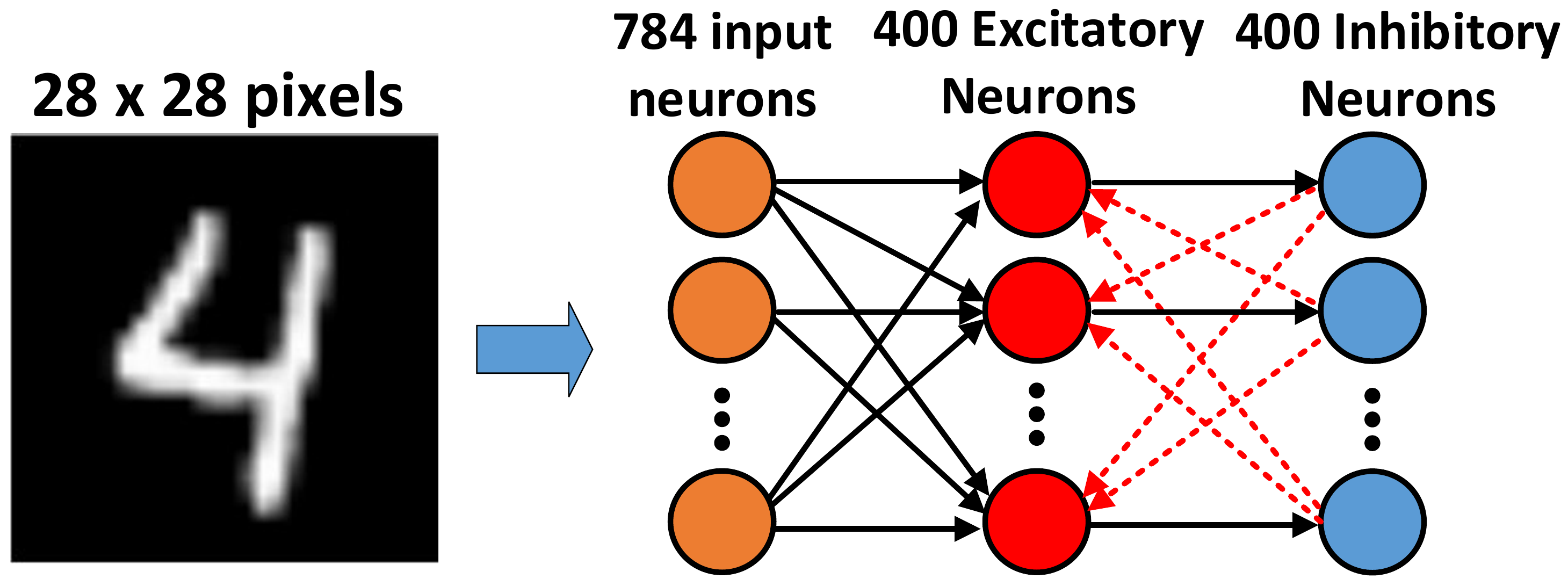}
	\caption{SNN architecture: input neuron layer encodes the 28$\times$28 pixel MNIST image as Poisson skyrmion trains. The input neuron layer and excitatory neuron layer are fully connected. The forward connection between excitatory- and inhibitory- neuron is one-to-one, while the backward connection is one-to-all, except the excitatory- and inhibitory- neuron has forward path.}
	\label{SNN}
\end{figure}

\textbf{\underline{System:}} In order to evaluate the performance of proposed AS-SNN, we have conducted the system level simulation through the spiking neural network simulator "BRIAN" \cite{goodman2009brian}, with MNIST \cite{lecun1998mnist} as the sample dataset. A simple three layer (i.e. input-, excitatory- and inhibitory- neuron layer) spiking neural network \cite{diehl2015unsupervised}, as shown in Fig.~\ref{SNN}, is used as the test structure. Our AS-SNN with binary synapse shows 87.1\% inference accuracy, which merely degrade 4.4\% in comparison to 91.5\% accuracy of SNN with analog synaptic weight. It is noteworthy that the inference accuracy can be further improved by increasing the number of excitatory neurons.

\bibliographystyle{ACM-Reference-Format}
\bibliography{sigproc} 

\end{document}